\def\year{2015}
\documentclass[letterpaper]{article}
\usepackage{aaai}
\usepackage{times}
\usepackage{helvet}
\usepackage{courier}
\usepackage{color}
\usepackage{url}
\usepackage{hyperref}
\usepackage{danudefs}
\usepackage{graphicx}

\usepackage{numprint}
\npdecimalsign{.}
\nprounddigits{4}

\usepackage{algorithmic}
\usepackage{algorithm}

\begin{document}
%
\title{Learning Word Representations from Relational Graphs}

\author{Danushka Bollegala \ \ Takanori Maehara  \ \ Yuichi Yoshida \ \ Ken-ichi Kawarabayashi\\
The University of Liverpool, Liverpool, L69 3BX, United Kingdom.  \\
National Institute of Informatics, 2-1-2, Hitotsubashi, Chiyoda-ku, Tokyo, 101-8430, Japan. \\
JST, ERATO, Kawarabayashi Large Graph Project. \\
Preferred Infrastructure, Inc., Hongo 2-40-1, Bunkyo-ku, Tokyo 113-0033, Japan.
}

\maketitle
\begin{abstract}
\begin{quote}
Attributes of words and relations between two words are central to numerous tasks in Artificial Intelligence such as
knowledge representation, similarity measurement, and analogy detection.
Often when two words share one or more attributes in common, they are connected by some semantic relations.
On the other hand, if there are numerous semantic relations between two words, we can expect some of
the attributes of one of the words to be inherited by the other. 
Motivated by this close connection between attributes and relations, given a relational graph in which words are inter-connected via
numerous semantic relations, we propose a method to learn a latent representation for the individual words.
The proposed method considers not only the co-occurrences of words as done by existing approaches for
word representation learning, but also the semantic relations in which two words co-occur.
To evaluate the accuracy of the word representations learnt using the proposed method,
we use the learnt word representations  to solve semantic word analogy problems.
Our experimental results show that it is possible to learn better word representations by using semantic semantics between words.
\end{quote}
\end{abstract}

\section{Introduction}
\label{sec:intro}

The notions of attributes and relations are central to Artificial Intelligence. In Knowledge Representation (KR)~\cite{KRbook},
a concept is described using its attributes and the relations it has with other concepts in a domain.
If we already know a particular concept such as \textit{pets}, we can describe a new concept such as \textit{dogs}
by stating the semantic relations that the new concept shares with the existing concepts such as \textit{dogs} \textsf{belongs-to} \textit{pets}.
Alternatively, we could describe a novel concept by listing all the attributes it shares with existing concepts.
In our example, we can describe the concept \textit{dog} by listing attributes such as \textit{mammal}, 
\textit{carnivorous}, and \textit{domestic animal} that it shares with another concept such as the \textit{cat}. 
Therefore, both attributes and relations can be considered as alternative descriptors of the same knowledge.
This close connection between attributes and relations can be seen in knowledge representation schemes such as predicate logic,
where attributes are modelled by predicates with a single argument whereas relations are modelled by predicates
with two or more arguments.

Learning representations of words is an important task with numerous applications~\cite{Bengio:IEEE:2013}.
Better representations of words can improve the performance in numerous natural language
processing tasks that require word representations such as language modelling~\cite{Collobert:2011,Bengio:JMLR:2003},  
part-of-speech tagging~\cite{zheng-chen-xu:2013:EMNLP}, sentiment classification~\cite{socher-EtAl:2011:EMNLP},
and dependency parsing~\cite{socher-EtAl:2013:ACL2013,Socher:ICML:2011}. 
For example, to classify a novel word into a set of existing categories one can measure the cosine
similarity between the words in each category and the novel word to be classified. Next, the novel word can be assigned to
the category of words that is most similar to~\cite{Huang:ACL:2012}. 
However, existing methods for learning word representations only consider the co-occurrences of two words within a 
short window of context, and ignore any semantic relations that exist between the two words.

Considering the close connection between attributes and relations,
an obvious question is, \textit{can we learn better word representations by considering the semantic relations that exist among words?}
More importantly whether word representations learnt by considering the semantic relations among words
could outperform methods that focus only on the co-occurrences of two words, ignoring the semantic relations.
We study these problems in this paper and arrive at the conclusion that it is indeed possible to learn better word representations
by considering the semantic relations between words.
Specifically, given as input a relational graph, a directed labelled weighted graph where vertices represent words and edges represent 
numerous semantic relations that exist between the corresponding words, 
we consider the problem of learning a vector representation for each vertex (word) in the graph
and a matrix representation for each label type (pattern).
The learnt word representations are evaluated for their accuracy by using them to solve semantic word analogy questions
on a benchmark dataset.

Our task of learning word attributes using relations between words is challenging because of several reasons.
First, there can be multiple semantic relations between two words. For example, consider the two words \textit{ostrich} and \textit{bird}.
An ostrich \textsf{is-a-large} bird as well as \textsf{is-a-flightless} bird. In this regard, the relation between the two words
\textit{ostrich} and \textit{bird} is similar to the relation between \textit{lion} and \textit{cat} (lion is a large cat) as well as to the relation
between \textit{penguin} and \textit{bird} (penguin is a flightless bird). Second, a single semantic relation can be expressed using multiple
lexical patterns. For example, the two lexical patterns \textbf{X} \textit{is a large} \textbf{Y} and \textit{large} \textbf{Y}s \textit{such as} \textbf{X}
represent the same semantic relation \textsf{is-a-large}. 
Beside lexical patterns, there are other representations of semantic relations between words
such as POS patterns and dependency patterns. An attribute learning method that operates on semantic relations must be able to
handle such complexities inherent in semantic relation representations. 
Third, the three-way co-occurrences between two words and a pattern describing a semantic relation is much sparser 
than the co-occurrences between two words (ignoring the semantic relations) or occurrences of individual words. 
This is particularly challenging for existing methods of representation learning where one must observe 
a sufficiently large number of co-occurrences to learn an accurate representation for words.

Given a \textit{relational graph}, a directed labeled weighted graph describing various semantic relations
 that exist between words denoted by the vertices of the graph,
we propose an unsupervised method to factorise the graph and assign latent attributional vectors $\vec{x}(u)$ for each vertex (word) $u$
and a matrix $\mat{G}(l)$ to each label (pattern) $l$ in the graph. Then, the co-occurrences between two words $u$ and $v$ in 
a pattern $l$ that expresses some semantic relation is modelled as the scalar product $\vec{x}(u)\T \mat{G}(l) \vec{x}(v)$.
A regression model that minimises the squared loss between the predicted co-occurrences according to the proposed method
and the actual co-occurrences in the corpus is learnt.
In the relational graphs we construct, 
the edge between two vertices is labeled using the patterns that co-occur with the corresponding words,
and the weight associated with an edge represents the strength of co-occurrence between the two words under the pattern.

Our proposed method does not assume any particular pattern extraction method or a co-occurrence weighting measure.
Consequently, the proposed method can be applied to a wide-range of relational graphs, both manually created ones such as
ontologies, as well as automatically extracted ones from unstructured texts.
For concreteness of the presentation, we consider relational graphs where the semantic relations between words are represented using
lexical patterns, part-of-speech (POS) patterns, or dependency patterns. 
Moreover, by adjusting the dimensionality of the decomposition, it is possible to obtain word representations at different granularities.
Our experimental results show that the proposed method obtain robust performances over a wide-range of relational graphs
constructed using different pattern types and co-occurrence weighting measures.
It learns compact and dense word representations with as low as 200 dimensions. 
Unlike most existing methods for word representation learning, our proposed method considers the semantic relations
that exist between two words in their co-occurring contexts.
To evaluate the proposed method, we use the learnt word representations 
to solve semantic analogy problems in a benchmark dataset~\cite{Milkov:2013}.

\section{Related Work}

Representing the semantics of a word is a fundamental step in many NLP tasks. 
Given word-level representations, numerous methods have been proposed in compositional semantics
 to construct phrase-level, sentence-level, or document-level 
 representations~\cite{Grefenstette:SEM:2013,socher-EtAl:2012:EMNLP-CoNLL}. Existing methods for creating word representations can be broadly categorised into two groups: \textit{counting-based} methods, and \textit{prediction-based} methods.

Counting-based approaches follow the distributional hypothesis \cite{Firth:1957}
which says that the meaning of a word can be represented using
the co-occurrences it has with other words. By aggregating the words that occur within a pre-defined window of context surrounding
all instances of a word from a corpus and by appropriately weighting the co-occurrences, it is possible to represent
the semantics of a word. 
Numerous definitions of co-occurrence such as within a proximity window or involved in a particular
dependency relation etc. and co-occurrence measures have been proposed in the literature \cite{Baroni:DM}. 
This counting-based bottom-up approach often results in sparse word representations. Dimensionality reduction techniques such as
the singular value decomposition (SVD) have been employed to overcome this problem in tasks such as
measuring similarity between words using the learnt word representations \cite{Turney:JAIR:2010}.

Prediction-based approaches for learning word representations model the problem of representation learning as
a prediction problem where the objective is to predict the presence (or absence) of a particular word in the context of another word.
Each word $w$ is assigned a feature vector $\vec{v}_{w}$ of fixed dimensionality such that the accuracy of predictions of other words
made using $\vec{v}_{w}$ is maximised. Different objective functions for measuring the prediction accuracy such as
perplexity or classification accuracy, and different optimisation methods have been proposed.
For example, Neural Network Language Model (NLMM) \cite{Bengio:JMLR:2003} learns word representations to 
minimise perplexity in a corpus.
The Continuous Bag-Of-Words model (CBOW)~\cite{Mikolov:NIPS:2013} uses the representations of all the words $c$ in the context 
of $w$ to predict the existence of $w$, whereas the skip-gram model~\cite{Mikolov:NAACL:2013,Milkov:2013}
learns the representation of a word $w$ by predicting the words $c$ that appear in the surrounding context of $w$.
Noise contrastive estimation has been used to speed-up the training 
of word occurrence probability models to learn word representations~\cite{Mnih:2013}.

Given two words $u$ and $v$ represented respectively by
vectors $\vec{x}(u)$ and $\vec{x}(v)$ of equal dimensions, 
GloVe~\cite{Pennington:EMNLP:2014} learns a linear regression model to 
minimise the squared loss between the inner-product $\vec{x}(u)\T\vec{x}(v)$ and the logarithm of the
co-occurrence frequency of $u$ and $v$. They show that this minimisation problem results in vector spaces
that demonstrate linear relationships observed in word analogy questions.
However, unlike our method, GloVe \textit{does not} consider the semantic relations that exist between two words
when they co-occur in a corpus. In particular, GloVe can be seen as a special case of our proposed method when
we replace all patterns by a single pattern that indicates co-occurrence, ignoring the semantic relations.
 
Our work in this paper can be categorised as a prediction-based method for word representation learning.
However, prior work in prediction-based word representation learning have been limited to considering the
co-occurrences between two words, ignoring any semantic relations that exist between the two words in their
co-occurring context. On the other hand, prior studies on counting-based approaches show 
that specific co-occurrences denoted by dependency relations are particularly useful for
creating semantic representations for words \cite{Baroni:DM}. 
Interestingly, prediction-based approaches have shown to outperform
counting-based approaches in comparable settings~\cite{baroni-dinu-kruszewski:2014:P14-1}.
Therefore, it is natural for us to consider the incorporation of
semantic relations between words into prediction-based word representation learning.
However, as explained in the previous section, three-way co-occurrences of two words and semantic relations
expressed by contextual patterns are problematic due to data sparseness. 
Therefore, it is non-trivial to extend the existing prediction-based word representation methods to 
three-way co-occurrences. 

Methods that use matrices to represent adjectives~\cite{Baroni:EMNLP:2010}
or relations~\cite{Socher:NIPS:2013b} have been proposed.
However, high dimensional representations are often difficult to learn because of their computational cost~\cite{paperno-pham-baroni:2014:P14-1}. 
Although we learn matrix representations for patterns as a byproduct, our final goal is the vector representations for words.
An interesting future research direction would be to investigate the possibilities of using the
learnt matrix representations for related tasks such as relation clustering~\cite{Duc:WI:2010}.

\section{Learning Word Representations from Relational Graphs}

\subsection{Relational Graphs}

We define a \textit{relational graph} $\cG(\cV, \cE)$ as a directed labelled weighted graph 
where the set of vertices $\cV$ denotes words in the vocabulary, and the set of edges $\cE$ denotes 
the co-occurrences between word-pairs and patterns. 
A pattern is a predicate of two arguments and expresses a semantic relation between the two words.
Formally, an edge $e \in \cE$ connecting two
vertices $u, v \in \cV$ in the relational graph $\cG$ is a tuple $(u,v, l(e), w(e))$,
where $l(e)$ denotes the label type corresponding to the pattern that co-occurs with the two words
$u$ and $v$ in some context, and $w(e)$ denotes the co-occurrence strength between $l(e)$ and
the word-pair $(u, v)$.
Each word in the vocabulary is represented by a unique vertex in the relational graph and each pattern is represented by a unique label type.
Because of this one-to-one correspondence between words and vertices, and patterns and labels, we will interchange those terms
in the subsequent discussions.
The direction of an edge $e$ is defined such that the first slot (ie. \textbf{X}) matches with $u$, and the second
slot (ie. \textbf{Y}) matches with $v$ in a pattern $l(e)$.
Note that multiple edges can exist between two vertices in a relational graph corresponding to different patterns.
Most manually created as well as automatically extracted ontologies can be represented as relational graphs.

\begin{figure}[t]
\centering
\includegraphics[width=60mm]{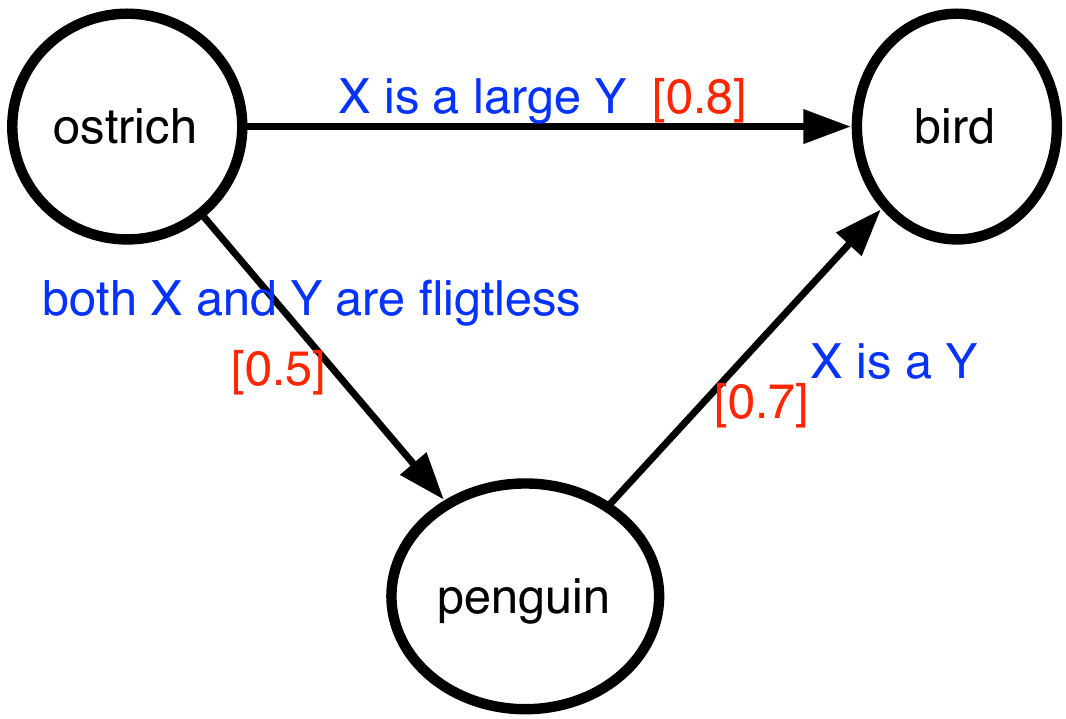}
\caption{A relational graph between three words.}
\label{fig:relgraph}
\vspace{-5mm}
\end{figure}

Consider the relational graph shown in Figure \ref{fig:relgraph}.
For example, let us assume that we observed the context \textit{ostrich is a large bird that lives in Africa} in a corpus.
Then, we extract the lexical pattern \textit{\textbf{X} is a large \textbf{Y}} between \textit{ostrich}
and \textit{bird} from this context and include it in the relational graph
by adding two vertices each for \textit{ostrich} and \textit{bird},
and an edge from \textit{ostrich} to  \textit{bird}.
Such lexical patterns have been used for related tasks such as measuring semantic similarity between words~\cite{Bollegala:JSAI:2007}.
The co-occurrence strength between a word-pair and a pattern can be 
computed using an association measure such as the positive pointwise mutual information (PPMI).
Likewise, observation of the contexts \textit{both ostrich and penguin are flightless birds} and
\textit{penguin is a bird} will result in the relational graph shown in Figure \ref{fig:relgraph}.

\subsection{Learning Word Representations}
\label{sec:attrfac}

Given a relational graph as the input, we learn $d$ dimensional vector representations for each vertex in the graph.
The dimensionality $d$ of the vector space is a pre-defined parameter of the method, and by adjusting it one can 
obtain word representations at different granularities. Let us consider two vertices $u$ and $v$ connected by an
edge with label $l$ and weight $w$. We represent the two words $u$ and $v$ respectively
by two vectors $\vec{x}(u), \vec{x}(v) \in \R^{d}$, and the label $l$ by a matrix $\mat{G}(l) \in \R^{d \times d}$.
We model the problem of learning optimal word representations $\hat{\vec{x}}(u)$ and
pattern representations $\hat{\mat{G}}(l)$
as the solution to the following squared loss minimisation problem

\begin{equation}
\small
\label{eq:objective}
\argmin_{\vec{x}(u) \in \R^{d}, \mat{G}(l) \in \R^{d \times d}} \frac{1}{2}\sum_{(u,v,l,w) \in \cE} 
{(\vec{x}(u)\T \mat{G}(l) \vec{x}(v) - w)}^{2} .
\end{equation}

The objective function given by Eq.~\ref{eq:objective} is jointly non-convex in 
both  word representations $\vec{x}(u)$ (or alternatively $\vec{x}(v)$)
and pattern representations $\mat{G}(l)$. However, if $\mat{G}(l)$ is positive semidefinite, 
and one of the two variables is held fixed, then the objective
function given by Eq.~\ref{eq:objective} becomes convex in the other variable.
This enables us to use Alternating Least Squares (ALS)~\cite{Boyd:ADMM:2010} method to solve the optimisation problem.
To derive the stochastic gradient descent (SGD) updates for the parameters in the model,
let us denote the squared loss associated with a single edge $e = (u,v,l,w)$ in the relational graph $\cG$ by $E(e)$,
given by,
\begin{equation}
\label{eq:per-instance}
E = \frac{1}{2} \sum_{(u,v,l,w) \in \cE}(\vec{x}(u)\T \mat{G}(l) \vec{x}(v) - w)^{2} . 
\end{equation}
In Eq. \ref{eq:per-instance}. The gradient of the error w.r.t. $\vec{x}(u)$ and $\mat{G}(l)$ are given by,
{\small
\begin{eqnarray}
\label{eq:grad:u}
\grad_{\vec{x}(u)}E &&=  \sum_{(u,v,l,w) \in \cE}(\vec{x}(u)\T \mat{G}(l) \vec{x}(v) - w) \mat{G}(l)\vec{x}(v) +  \nonumber \\ 
				  &&  \sum_{(v,u,l,w) \in \cE}(\vec{x}(v)\T \mat{G}(l) \vec{x}(u) - w) \mat{G}(l)\T \vec{x}(u)
\end{eqnarray}}
\begin{equation}
\small
\grad_{\mat{G}(l)}E = \sum_{(u,v,l,w) \in \cE}(\vec{x}(u)\T \mat{G}(l) \vec{x}(v) - w) \vec{x}(v) \vec{x}(u)\T 
\label{eq:grad:G}
\end{equation}
In Eq. \ref{eq:grad:G}, $\vec{x}(v)\vec{x}(u)\T$ denotes the outer-product between the two vectors $\vec{x}(v)$ and
$\vec{x}(u)$, which results in a $d \times d$ matrix.
Note that the summation in Eq.~\ref{eq:grad:u}  is taken over the edges that contain either
$u$ as a start or an end point, and the summation in Eq.~\ref{eq:grad:G} is taken over the edges that 
contain the label $l$.

The SGD update for the $i$-th dimension of $\vec{x}(u)$ is given by,
\begin{equation}
\small
\label{eq:update:u}
\vec{x}(u)_{(i)}^{(t+1)} = \vec{x}(u)_{(i)}^{(t)} - \frac{\eta_{0}}{\sqrt{\sum\limits_{t'=1}^{t}
 {\left( \grad_{{\vec{x}(u)}^{(t')}}E \right)_{(i)}}^{2}}}  \left( \grad_{{\vec{x}(u)}^{(t)}}E \right)_{(i)} .
\end{equation}
Here, $\left( \grad_{{\vec{x}(u)}^{(t)}}E \right)_{(i)}$ denotes the $i$-th dimension of the gradient vector of $E$ w.r.t. ${\vec{x}(u)}^{(t)}$,
and the superscripts $(t)$ and $(t+1)$ denote respectively the current and the updated values.
We use adaptive subgradient method (AdaGrad) \cite{Duchi:JMLR:2011} 
to schedule the learning rate. The initial learning rate, $\eta_{0}$ is set to $0.0001$ in our experiments.

Likewise, the SGD update for the $(i,j)$ element of $\mat{G}(l)$ is given by,
\begin{equation}
\small
\mat{G}(l)_{(i,j)}^{(t+1)} = \mat{G}(l)_{(i,j)}^{(t)} - \frac{\eta_{0}}{\sqrt{\sum\limits_{t'=1}^{t} 
{\left( \grad_{{\mat{G}(l)}^{(t')}}E \right)_{(i,j)}}^{2}}}  \left( \grad_{{\mat{G}(l)}^{(t)}}E \right)_{(i,j)} .
\label{eq:update:G}
\end{equation}

Recall that the positive semidefiniteness of $\mat{G}(l)$ is a requirement for the convergence of the above procedure.
For example, if $\mat{G}(l)$ is constrained to be diagonal then this requirement can be trivially met.
However, doing so implies that $\vec{x}(u)\T \mat{G}(l) \vec{x}(v) = \vec{x}(v)\T \mat{G}(l) \vec{x}(u)$, which
means we can no longer capture asymmetric semantic relations in the model.
Alternatively, without constraining $\mat{G}(l)$ to diagonal matrices,
we numerically guarantee the positive semidefiniteness of $\mat{G}(l)$ by adding a small noise term $\delta \mat{I}$
after each update to $\mat{G}(l)$, where $\mat{I}$ is the $d \times d$ identity matrix and $\delta \in R$ is a small perturbation coefficient,
which we set to $0.001$ in our experiments.

Pseudo code for the our word representation learning algorithm is shown in Algorithm \ref{algo:opt}.
Algorithm \ref{algo:opt} initialises word and pattern representations randomly by sampling from the zero-mean
and unit variance Gaussian distribution. Next, SGD updates are performed alternating between updates for $\vec{x}(u)$
and $\mat{G}(l)$ until a pre-defined number of maximum epochs is reached. 
Finally, the final values of $\vec{x}(u)$ and $\mat{G}(l)$ are returned.

\begin{algorithm}[t]       
\small
\caption{Learning word representations.}        
\label{algo:opt}                         
\begin{algorithmic}[1]         
\REQUIRE Relational graph $\cG$, dimensionality $d$ of the word representations, maximum epochs $T$,
initial learning rate $\eta_{0}$.
\ENSURE Word representations $\vec{x}(u)$ for words $u \in \cV$.
\medskip
\STATE \textbf{Initialisation:} For each vertex (word) $u \in \cV$, randomly sample $d$ dimensional real-valued vectors from the normal distribution. 
For each label (pattern) $l \in \cL$, randomly sample $d \times d$ dimensional real-valued matrices from the normal distribution. \label{line:one}
\FOR {$t = 1, \ldots, T$} 
	\FOR {edge $(u,v,l,w) \in \cG$}
		\STATE Update $\vec{x}(u)$ according to Eq. \ref{eq:update:u}
		\STATE Update $\mat{G}(l)$ according to Eq. \ref{eq:update:G}
	 \ENDFOR
\ENDFOR
\RETURN $\vec{x}(u), \mat{G}(l)$ 
\end{algorithmic}
\end{algorithm}

\section{Experiments}

\subsection{Creating Relational Graphs}

We use the English ukWaC\footnote{\url{http://wacky.sslmit.unibo.it/doku.php?id=corpora}}
corpus in our experiments.
ukWaC is a 2 billion token corpus constructed from the Web limiting the crawl to \textbf{.uk} domain and medium-frequency
words from the British National Corpus (BNC). The corpus is lemmatised and Part-Of-Speech tagged using the 
TreeTagger\footnote{\url{www.cis.uni-muenchen.de/~schmid/tools/TreeTagger}}.
Moreover, MaltParser\footnote{\url{www.maltparser.org}} is used to create a dependency parsed version of the ukWaC corpus. 

To create relational graphs, we first compute the co-occurrences of words in sentences in the ukWaC corpus.
For two words $u$ and $v$ that co-occur in more than $100$ sentences, we create two word-pairs $(u,v)$ and $(v,u)$.
Considering the scale of the ukWaC corpus, low co-occurring words often represents misspellings or non-English terms.

Next, for each generated word-pair, we retrieve the set of sentences in which the two words co-occur.
For explanation purposes let us assume that the two words $u$ and $v$ co-occur in a sentence $s$.
We replace the occurrences of $u$ in $s$ by a slot marker \textbf{X} and $v$ by \textbf{Y}.
If there are multiple occurrences of $u$ or $v$ in $s$, we select the closest occurrences 
$u$ and $v$ measured by the number of tokens that appear in between the occurrences in $s$.
Finally, we generate lexical patterns by limiting prefix (the tokens that appears before \textbf{X} in $s$),
midfix (the tokens that appear in between \textbf{X} and \textbf{Y} in $s$), and suffix
(the tokens that appear after \textbf{Y} in $s$) each separately to a maximum length of $3$ tokens. For example,
given the sentence \textit{ostrich is a large bird that lives in Africa}, we will extract the lexical patterns
\textbf{X} \textit{is a large} \textbf{Y}, \textbf{X} \textit{is a large} \textbf{Y} \textit{that}, 
\textbf{X} \textit{is a large} \textbf{Y} \textit{that lives},
and \textbf{X} \textit{is a large} \textbf{Y} \textit{that lives in}.
We select lexical patterns that co-occur with at least two word pairs for creating a relational graph.

In addition to lexical patterns, we generate POS patterns by replacing each lemma in a lexical pattern by its POS tag.
POS patterns can be considered as an abstraction of the lexical patterns. 
Both lexical and POS patterns are unable to capture semantic relations between two words if those words
are located beyond the extraction window. One solution to this problem is to use the dependency path between the two words
along the dependency tree for the sentence. We consider pairs of words that have a dependency relation between them in a sentence and
extract dependency patterns such as \textbf{X} \textit{direct-object-of} \textbf{Y}. Unlike lexical or POS patterns that
are proximity-based, the dependency patterns are extracted from the entire sentence. 

We create three types of relational graphs using $127,402$ lexical (\textbf{LEX}) patterns , 
$57,494$ POS (\textbf{POS}) patterns, and $6835$ dependency (\textbf{DEP}) patterns as edge labels.
Moreover, we consider several popular methods for weighting the co-occurrences between a word-pair and a pattern as follows.
\begin{description}
\item \textbf{RAW}: The total number of sentences in which a pattern $l$ co-occurs with a word-pair $(u,v)$ is considered as 
the co-occurrence strength $w$.
\item \textbf{PPMI}: The positive pointwise mutual information between a pattern $l$ and a word-pair $(u,v)$ computed as,
\[ \max \left(0, \log \left(\frac{h(u,v,l) \times h(*,*,*)}{h(u,v,*) \times h(*, *, l)} \right) \right) . \] Here, $h(u,v,l)$ denotes the total number of sentences
in which $u$, $v$, and $l$ co-occurs. The operator $*$ denotes the summation of $h$ over the corresponding variables.
\item \textbf{LMI}: The local mutual information between a pattern $l$ and a word pair $(u,v)$ is computed as,
\[ \frac{h(u,v,l)}{h(*,*,*)} \log \left(\frac{h(u,v,l) \times h(*,*,*)}{h(u,v,*) \times h(*, *, l)} \right) . \]
\item \textbf{LOG}: This method considers the logarithm of the raw co-occurrence frequency as the co-occurrence strength.
It has been shown to produce vector spaces that demonstrate vector substraction-based analogy representations \cite{Pennington:EMNLP:2014}.
\item \textbf{ENT}: Patterns that co-occur with many word-pairs tend to be generic ones that does not express a specific semantic relation.
Turney~\shortcite{Turney_CL} proposed the entropy of a pattern over word-pairs as a measure to down-weight the effect of such patterns. 
We use this entropy-based co-occurrence weighting method to weigh the edges in relational graphs.
\end{description}

\subsection{Evaluation}

We use the semantic word analogy dataset first proposed by Mikolov et al.~\shortcite{Milkov:2013}
and has been used in much previous work for evaluating word representation methods.
Unlike syntactic word analogies such as the past tense or plural forms of verbs which can be 
accurately captured using rule-based methods \cite{Lepage:ACL:2000}, 
semantic analogies are more difficult to detect using surface-level transformations.
Therefore, we consider it is appropriate to evaluate word representation methods using semantic word analogies.
The dataset contains $8869$ word-pairs that represent word analogies covering various semantic relations
 such as the capital of a country (e.g. \textit{Tokyo, Japan} vs. \textit{Paris, France}), 
 and family (gender) relationships (e.g. \textit{boy, girl} vs. \textit{king, queen}).

A word representation method is evaluated by its ability to correctly answer word analogy questions
using the word representations created by that method.
For example, the semantic analogy dataset contains word pairs such as 
(\textit{man}, \text{woman}) and (\textit{king}, \textit{queen}), where the semantic relations between the two words in the first pair
is similar to that in the second. Denoting the representation of a word $w$ by a vector $\vec{v}(w)$, we rank all words $w$ 
in the vocabulary according to their cosine similarities with the vector, $\vec{v}(king) - \vec{v}(man) + \vec{v}(woman)$.
The prediction is considered correct in this example only if the top most similar vector is $\vec{v}(queen)$.
During evaluations, we limit the evaluation to analogy questions where word representations have been learnt
for all the four words. Moreover, we remove three words that appear in the question from the set of candidate answers.
The set of candidates for a question is therefore the set consisting of fourth words in all semantic analogy questions
considered as valid (after the removal step described above), minus the three words in the question under consideration.
The percentage of correctly answered semantic analogy questions out of the total number of questions
in the dataset (ie. micro-averaged accuracy) is used as the evaluation measure.

\subsection{Results}
\label{sec:results}

To evaluate the performance of the proposed method on relational graphs created using
different pattern types and co-occurrence measures, we train $200$ dimensional 
word representations ($d = 200$) using Algorithm~\ref{algo:opt}.
$100$ iterations ($T = 100$) was sufficient to obtain convergence in all our experiments.
We then used the learnt word representations to obtain the accuracy values
shown in Table~\ref{tbl:patterns}. We see that the proposed method obtains
similar results with all pattern types and co-occurrence measures. This result shows 
the robustness of our method against a wide-range of typical methods for constructing relational graphs from unstructured texts.
For the remainder of the experiments described in the paper, we use the RAW co-occurrence frequencies as the
co-occurrence strength due to its simplicity.

\begin{table}[t]
\small
\centering
\caption{Semantic word analogy detection accuracy using word representations learnt by the proposed method from
relational graphs with different pattern types and weighting measures.}
\label{tbl:patterns}
\begin{tabular}{|l||c|c|c|} \hline
Measure		&	LEX		&	POS			&	DEP 		\\ \hline \hline
RAW		&	$26.61$	&	$25.35$		&	$24.68$		\\
PPMI		&	$25.00$	&	$25.51$		&	$24.48$		\\
LMI			&	$27.15$	&	$25.19$		&	$24.41$		\\
LOG			&	$23.01$	&	$25.55$		&	$24.47$		\\
ENT			&	$22.39$	&	$25.55$		&	$24.52$		\\ \hline
\end{tabular}
\vspace{-3mm}
\end{table}

\begin{figure}
\centering
\includegraphics[width=70mm]{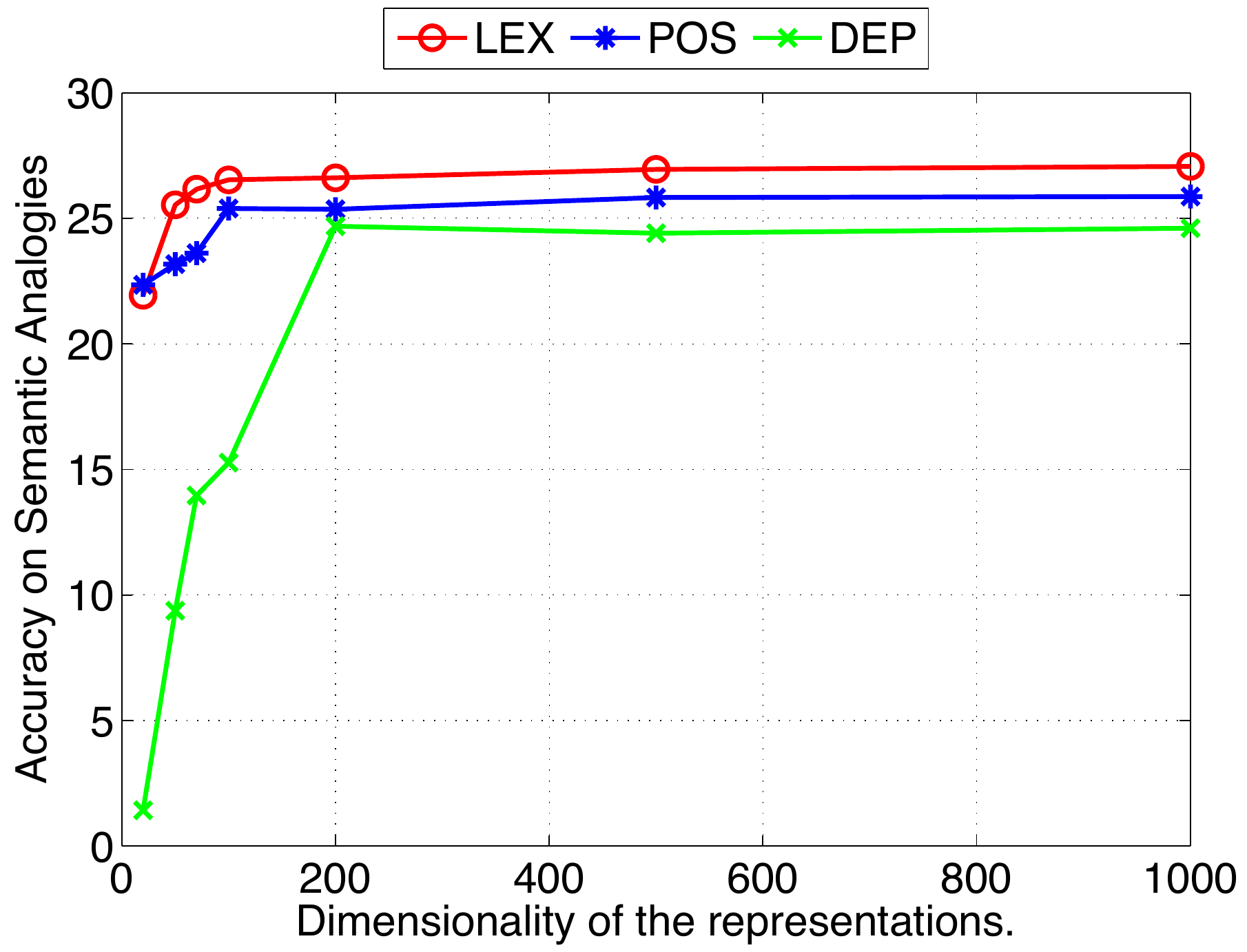}
\caption{The effect of the dimensionality of the word representation on the accuracy of the semantic analogy task.}
\label{fig:rankplot}
\vspace{-4mm}
\end{figure}

\begin{table*}[t]
\small
\centering
\caption{Comparison of the proposed method (denoted by \textbf{Prop}) against prior work on word representation
learning.} 
\label{tbl:overall}
\begin{tabular}{|l||c|c|c|c|c||c|}\hline
Method		&	capital-common	&	capital-world		&	city-in-state	& family (gender) 	& currency	&	Overall Accuracy\\ \hline \hline
SVD+LEX 	&	11.43			&	5.43				&	0			&	9.52			& 	0		&	3.84				\\ 
SVD+POS 	&	4.57				&	9.06				&	0			&	29.05		& 	 0		&	6.57				\\ 
SVD+DEP 	&	5.88				&	3.02				&	0			&	0			& 	0		&	1.11				 \\ 
CBOW 		&	8.49				&	5.26				&	4.95			&	47.82		& 2.37		&	10.58 		\\ 
skip-gram 	& 	9.15				&	9.34				&	5.97			&	\textbf{67.98}	& 5.29	&	14.86 		\\ 
GloVe		&	4.24				&	4.93				&	4.35			&	65.41		&	0		&	11.89				\\
Prop+LEX 	&	\textbf{22.87}	&	\textbf{31.42}	&	\textbf{15.83}&	61.19		& 25.0		& \textbf{26.61} \\ 
Prop+POS	&	22.55			&	30.82			&	14.98		&	60.48	       & 20.0		&	25.35 \\ 
Prop+DEP	&	20.92			&	31.40			&	15.27		&	56.19	       & 20.0		& 24.68 \\ \hline
\end{tabular}
\vspace{-3mm}
\end{table*}

To study the effect of the dimensionality $d$ of the representation that we learn on the accuracy of the semantic analogy task,
we plot the accuracy obtained using LEX, POS, and DEP relational graphs against the dimensionality of the representation
as shown in Figure~\ref{fig:rankplot}. We see that the accuracy steadily increases with the
dimensionality of the representation up to 200 dimensions and then becomes relatively stable. 
This result suggests that it is sufficient to use $200$ dimensions for all three types of relational graphs that we constructed.
Interestingly, among the three pattern types, LEX stabilises with the least number of dimensions followed by POS and DEP.
Note that LEX patterns have the greatest level of specificity compared to POS and DEP patterns, which abstract the 
surface-level lexical properties of the semantic relations. 
Moreover, relational graphs created with LEX patterns have the largest number of labels (patterns)
followed by that with POS and DEP patterns.
The ability of the proposed method to obtain better performances even with a highly specified, sparse and high-dimensional
feature representations such as the LEX patterns is important when applying the proposed method to large relational graphs.

We compare the proposed method against several word representation methods in Table~\ref{tbl:overall}. 
All methods in Table~\ref{tbl:overall} use $200$ dimensional vectors to represent a word.
A baseline method is created that shows the level of performance we can reach if we represent each word $u$ as a vector
of patterns $l$ in which $u$ occurs. First, we create a co-occurrence matrix between words $u$ and patterns $l$,
and use Singular Value Decomposition (SVD) to create $200$ dimensional projections for the words.
Because patterns represent contexts in which words appear in the corpus, this baseline
can be seen as a version of the Latent Semantic Analysis (LSA), that has been widely used to represent words and documents
in information retrieval. Moreover, SVD reduces the data sparseness in raw co-occurrences.
We create three versions of this baseline denoted by \textbf{SVD+LEX}, \textbf{SVD+POS}, and \textbf{SVD+DEP}
corresponding to relational graphs created using respectively LEX, POS, and DEP patterns.
\textbf{CBOW}~\cite{Mikolov:NIPS:2013}, \textbf{skip-gram}~\cite{Mikolov:NAACL:2013},
and \textbf{GloVe}~\cite{Pennington:EMNLP:2014} are previously proposed word representation learning methods.
In particular, \textbf{skip-gram} and \textbf{GloVe} are considered the current state-of-the-art methods.
We learn $200$ dimensional word representations using their original implementations with the default settings.
We used the same set of sentences as used by the proposed method to train these methods.
Proposed method is trained using $200$ dimensions and with three relational graphs
(denoted by \textbf{Prop+LEX}, \textbf{Prop+POS}, and \textbf{Prop+DEP}), weighted by RAW co-occurrences. 

From Table~\ref{tbl:overall}, we see that \textbf{Prop+LEX} 
obtains the best overall results among all the methods compared. 
We note that the previously published results 
for skip-gram and CBOW methods are obtained using a 100B token news corpus,
which is significantly large than the 2B token ukWaC corpus used in our experiments.
However, the differences among the \textbf{Prop+LEX}, \textbf{Prop+POS}, and \textbf{Prop+DEP} 
methods are not significantly different according to the Binomial exact test.
SVD-based baseline methods perform poorly indicating that de-coupling the 3-way co-occurrences between $(u,v)$ and
$l$ into 2-way co-occurrences between $u$ and $l$ is inadequate to capture the semantic relations between words.
\textbf{Prop+LEX} reports the best results for all semantic relations except for the \textit{family} relation.
Comparatively, higher accuracies are reported for the family relation by all methods,
whereas relations that involve named-entities such as locations are difficult to process.
Multiple relations can exist between two locations, which makes the analogy detection task hard.

\section{Conclusion}
We proposed a method that considers not only the co-occurrences of two words but also the semantic
relations in which they co-occur to learn word representations. 
It can be applied to manually created
relational graphs such as ontologies, as well as automatically extracted relational graphs from text corpora.
We used the proposed method to learn word representations from three types of relational graphs.
We used the learnt word representations to answer semantic word analogy questions
using a previously proposed dataset. 
Our experimental results show that lexical patterns are particularly useful for learning good word representations,
outperforming several baseline methods.
We hope our work will inspire future research in word representation learning to exploit the rich semantic relations that
exist between words, extending beyond simple co-occurrences.

\bibliographystyle{named}
\bibliography{rel2att}

\end{document}